\documentclass[10pt,twocolumn,letterpaper]{article}

\usepackage{cvpr}
\usepackage{times}
\usepackage{epsfig}
\usepackage{amsmath}
\usepackage{amssymb}
\usepackage{graphicx,booktabs,array}
\usepackage[utf8]{inputenc}
\usepackage{makecell}

\usepackage[pagebackref=true,breaklinks=true,letterpaper=true,colorlinks,bookmarks=false]{hyperref}

\cvprfinalcopy 

\def\cvprPaperID{****} 

\begin{document}

\title{RecipeSnap - a lightweight image to recipe model}

\author{Jianfa Chen\\
Georgia Institute of Technology\\
225 North Avenue, Atlanta, GA\\
{\tt\small jchen3105@gatech.edu}
\and
Yue Yin\\
Georgia Institute of Technology\\
225 North Avenue, Atlanta, GA\\
{\tt\small yyin79@gatech.edu}   
\and
Yifan Xu\\
Georgia Institute of Technology\\
225 North Avenue, Atlanta, GA\\
{\tt\small yxu349@gatech.edu}
}

\maketitle

\begin{abstract}

   In this paper we want to address the problem of automation for recognition of photographed cooking dishes and generating the corresponding food recipes. Current image-to-recipe models are computation expensive and require powerful GPUs for model training and implementation. High computational cost prevents those existing models from being deployed on portable devices, like smart phones. To solve this issue we introduce a lightweight image-to-recipe prediction model, RecipeSnap, that reduces memory cost and computational cost by more than 90\% while still achieving 2.0 MedR, which is in line with the state-of-the-art model. A pre-trained recipe encoder from \cite{DBLP:conf/cvpr/SalvadorGBD21} was used to compute recipe embeddings. Recipes from Recipe1M \cite{DBLP:journals/pami/MarinBOHSAWT21} dataset and corresponding recipe embeddings are collected as a recipe library (Figure \ref{tab:build recipe library}), which are used for image encoder training (Figure \ref{tab:image encoder training}) and image query (Figure \ref{tab:predict recipe from image}) later. We use MobileNet-V2 as image encoder backbone, which makes our model suitable to portable devices. This model can be further developed into an application for smart phones with a few effort. A comparison of the performance between this lightweight model to other heavy models are presented in this paper. Code, data and models are publicly accessible \footnote{https://github.com/jianfa/RecipeSnap-a-lightweight-image-to-recipe-model.git}.

\end{abstract}

\begin{figure}[ht] 
\centering
\includegraphics[width=0.5\textwidth]{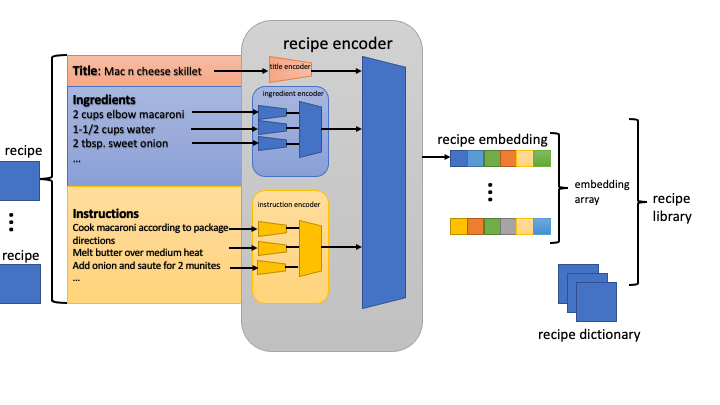}
\caption{Build recipe library. A pre-trained recipe encoder from \cite{DBLP:conf/cvpr/SalvadorGBD21} is used to compute recipe embeddings. Recipes from Recipe1M \cite{DBLP:journals/pami/MarinBOHSAWT21} dataset and corresponding recipe embeddings are collected as a recipe library.}
\label{tab:build recipe library}
\end{figure}

\begin{figure}[ht] 
\centering
\includegraphics[width=0.5\textwidth]{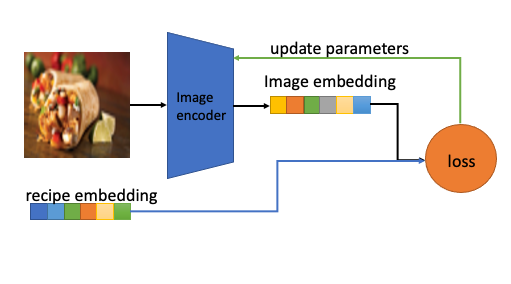}
\caption{Image encoder training. We use MobileNet V2 as image encoder backbone. Images from Recipe1M training partition combined with corresponding pre-computed recipe embeddings in recipe library are serving as input for model training. The loss function is same as the pair loss function in \cite{DBLP:conf/cvpr/SalvadorGBD21}. }
\label{tab:image encoder training}
\end{figure}

\begin{figure}[ht] 
\centering
\includegraphics[width=0.5\textwidth]{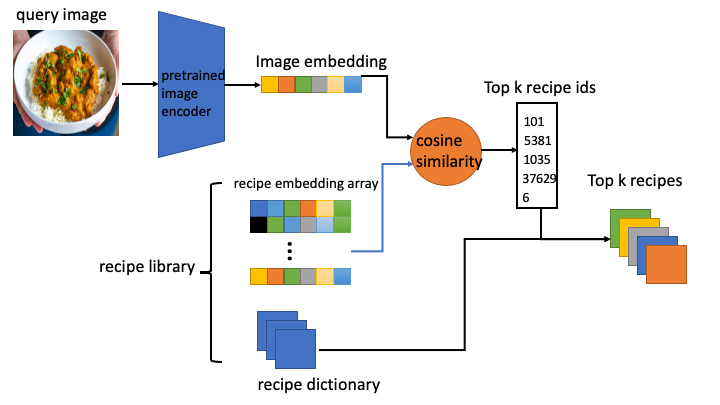}
\caption{Predict recipe from image. A query image is passed to pre-trained image encoder to get image embedding. Then we compute the cosine similarity between image embedding and recipe embedding array (numpy array). Only top k (here we use k=5) recipe embedding ids are collected. Corresponding top k recipes can be easily retrieved from recipe dictionary with embedding ids.  }
\label{tab:predict recipe from image}
\end{figure}

\section{Introduction}

\subsection{Introduction and Background}

With the rapid spread and easy access of social networks nowadays, countless pictures for cooking dishes and recipes are created and shared with everyone. As almost nothing in the world other than food has such a profound and tremendous impact in people's life this has created urging needs for machine learning algorithms to automatically retrieve associated recipes given input images of food or dishes. The mass quantity of user submitted cooking recipes and food images online brings the possibility of training networks to find digital cooking recipes (ingredient lists, cooking instructions etc.) relevant to the specific food image. Although it is also possible to reverse the image-to-recipe procedure that is to find relevant food images given recipes such tasks are unlikely to occur in real world. Usually people would search food images with keyword or recipe title in a searching engine.

After searching for and studying on some of the recent published works, we found that there are a couple of existing frameworks established in this particular fields. More details are illustrated in the following Section \ref{relatedwork}. However, all of them require large memory and powerful computation resource, like GPUs. Heavy model size prevent these models from deploying on mobile devices. After conducting a comprehensive analysis quantitatively and qualitatively, we decided to build a lightweight cross-modal framework that is suitable for mobile devices. 
\subsection{Related Work}
\label{relatedwork}

To start our work in the development of the lightweight prediction model to handle image-to-recipe retrieve tasks, we reviewed a couple of relevant papers. We mainly built our work upon the study of a couple of recent published papers that utilize multi-model framework to solve similar problems and here provided a brief summary of their work.

In \cite{DBLP:journals/pami/MarinBOHSAWT21} the authors use bi-directional LSTM model in the design of the two major components, ingredients and instructions, of recipe encoder. The two-stage structure has taken consideration of both forward and backward orderings and better suits the recipe representation than a simple LSTM model. For the image encoding authors adopt two major deep convolutional networks that has achieved proven record of success, VGG-16 and ResNet-50 models. 

In \cite{DBLP:conf/iclr/DosovitskiyB0WZ21} the authors got inspired by the success of Transformers in NLP field and conducted experiments on using standard Transformer model directly on image inputs. To realize their goal with minimum modification the authors processed the original images into patches and a sequence of linear embedding resulted is input into the Transformer. According to the experiment results provided from the paper, this model only achieved better performance when trained on larger size of dataset which range from 14M to 300M images. Among all that their Vision Transformer (ViT) produced excellent results when pre-trained on sufficient amount of data-points and transferred to smaller-scale tasks. 

Amaia et al. introduced a hierarchical Transformer model which encodes each recipe components eg. titles, ingredients and instructions individually \cite{DBLP:journals/pami/MarinBOHSAWT21}. In addition, the author proposed a self-supervised loss function computed based on pairs of the individual recipe components to leverage the semantic relationships within recipes. Whereas in \cite{DBLP:conf/cvpr/SalvadorGBD21} the author developed a neural with joint embedding learned on the recipes and images in common space. In the model a high-level classification task was added to further improve the classification performance. \cite{DBLP:journals/corr/abs-1812-06164}

Hao and others proposed a end-to-end trainable Adversial Cross-Model Embedding (ACME) validated using the Recipe1M dataset \cite{DBLP:conf/cvpr/WangSLLH19} to resolve the food retrieval task where their specific goal is to develop a constructive mapping space from recipes to matching food images. As shown in the paper, the proposed network achieved out-performance over all baseline models. Specifically, the authors proposed an refined triplet loss methodology which increased the convergence and accuracy of the neural network. In addition, an adversarial loss scheme was utilized to align the distribution of the encoded features from recipes and food images. 

We noticed that these papers use more and more complex neural networks to achieve better performance. In the meanwhile, prerequisite on hardware, like memory and computation power, is being more and more difficult to meet. Given the prevalence of portable devices, like smart phones and tablets, we believe it will be greatly advanced if these models can be deployed on portable devices. To build a lightweight image-to-recipe model that can be easily trained and deployed on smart phones, we also reviewed some existing models that are suitable for smart phones.


 Mark et al. introduced the MobileNet V2 model in \cite{DBLP:journals/corr/HowardZCKWWAA17} which is a mobile framework based on  a inverted residual structure. This model is specifically designed to provide a convenient and efficient network for actual usage. The approaches that's been taken by the author to meet their goal including: within the structure a shortcut path are built to connect the bottleneck layers; they use depth-wise convolution filters to extract features; they remove non-linearity from narrow layers to obtain better representing power. Additionally the paper described valid ways of application of this mobile network framework to object detection. Model size and complexity comparison is disclosed in Table \ref{tab:model comparision table}.

\subsection{Dataset}

Our model was training on the Recipe1M dataset, which is a large dataset contains paired data points - cooking recipes and dish images. This dataset was first introduced and collected by Javier et al. \cite{DBLP:conf/cvpr/WangSLLH19} by scraping from a variety of popular cooking websites. The overall dataset contains over one million cooking recipes and about 900K food images. As shown in Table \ref{tab:table1} the data has been broken down in a $70\%$,$15\%$,$15\%$ manner for the purposes of training/validation/testing of the model. 

The cooking recipes and linked images are initially scraped from over two dozen widely used cooking websites and handled via a pipeline process. Only related content is generated from the raw HTML that is to say unnecessary blank space, non-ASCII characters etc. have been left out during the extraction process. After removing the duplicates or close-matches, the final dataset is consolidated into a compressed JSON schema and ready for use. 

Due to the limited time and resource we have for this project, we believe the use of the Recipe1M dataset should fulfill our goal given the following considerations:
\begin{itemize}
    \item This dataset has over one million recipe and image data which is among the largest publicly available datasets ever been used in related work.
    \item Recipe1M contains paired recipe and image data points which well suits the purpose of training a multi-model framework.
    \item The dataset is clean and consolidated with any duplicates and near-matches taken out which save us time for more model experiments.

\end{itemize}


\begin{table}[h!]
  \begin{center}
    \caption{Dataset Sizes}
    \label{tab:table1}
    \begin{tabular}{l|c|c} 
      \textbf{Partition} & \textbf{Num of Recipes} & \textbf{Num of Images}\\
      \hline
      Training & 720,639 & 619,508\\
      Validation & 155,036 & 133,860\\
      Testing & 154,045 & 134,338\\
      \hline
      Total & 1,029,720 & 887,706\\

    \end{tabular}
  \end{center}
\end{table}



\begin{table}[h!]
  \begin{center}
    \caption{Comparison of model size and complexity}
    \label{tab:model comparision table}
    \begin{tabular}{l|c|c} 
      \textbf{Model} & \textbf{Param.} & \textbf{FLOPs}\\
      \hline
      AlexNet\cite{li2018harmonious} & 58.3M & 725M\\
      VGG16\cite{li2018harmonious} & 134.2M & 15.5G\\
      ResNet50\cite{li2018harmonious} & 23.5M & 3.80G\\
      GoogLeNet\cite{li2018harmonious} & 6.0M & 1.57G\\
      ViT(base)\cite{dang2021revisiting} & 86M & 15.85G\\
      MobileNetV2\cite{sandler2018mobilenetv2} & 3.34M & 319M\\
      \hline

    \end{tabular}
  \end{center}
\end{table}

\begin{table*}[t]
\caption{Comparison between two SOTA models}
\label{tab:SOTA_performance}
\begin{tabular}{ccccccc}

    \hline
    {Model}&\multicolumn{4}{c}{Performance}&\multicolumn{2}{c}{Cost (one NVIDIA K80 GPU)}\\
    \cline{2-5}\cline{6-7}\\
         & MedR & Recall@1 & Recall @5 & Recall @10 & Avg. Training Time per Epoch & Epoch \\
    LSTM \cite{DBLP:journals/pami/MarinBOHSAWT21} &  5.1    & 0.24      &  0.52        &   0.64        &        -                     &   -   \\
    Hierarchical Transformer \cite{DBLP:conf/cvpr/SalvadorGBD21} & 1.0     & 0.60       & 0.876         &   0.929        &       -                      & -    \\
    LSTM (ours) &  14.15    & 0.1165      &  0.3149        &   0.4409        &        ~ 2.5 hours                     &   20   \\
    Hierarchical Transformer (ours) & 16.20     & 0.0969       & 0.2870         &   0.4063        &       ~9 hours                      & 2    \\
    MobileNet-v2 + Transformer (ours) & 2     & 0.4123       & 0.7187         &   0.8210        &       ~3.5 hours                      & 3    \\

    \hline
\end{tabular}
\label{tab:multicol}
\end{table*}

\section{Approach}
\label{s_approach}
Our ultimate goal is to build an application that can carry out the image-to-recipe tasks on standard mobile devices. Therefore, we naturally face the trade-off between a deeper and more complicated network model with higher accuracy vs. a less-powered model with lighter weight and low latency. To understand the trade-off better, we have explored and compared a few different approaches. \\
We first start with the models that deliver the state-of-the-art (SOTA) performance on accuracy. Specifically, we focus on two models that frame the problem as a cross-modal recipe retrieval task\cite{DBLP:journals/pami/MarinBOHSAWT21}\cite{DBLP:conf/cvpr/SalvadorGBD21}. The main difference between these two models lies in the design of recipe encoder: 1) one uses a two-stage LSTM \cite{DBLP:journals/pami/MarinBOHSAWT21} while the other uses hierarchical transformers \cite{DBLP:journals/pami/MarinBOHSAWT21}. 2) \cite{DBLP:journals/pami/MarinBOHSAWT21} introduces an auxiliary self-supervised task and loss to learn the semantic relationship between recipe components, strengthening the recipe encoder. For image encoder, \cite{DBLP:journals/pami/MarinBOHSAWT21} uses the visual transformer ViT \cite{DBLP:conf/iclr/DosovitskiyB0WZ21} while \cite{DBLP:conf/cvpr/SalvadorGBD21} uses the ResNet-50 \cite{DBLP:conf/cvpr/HeZRS16}.  \\
We train these two models from scratch to compare the performance and cost. For evaluation, we follow the procedure described in \cite{DBLP:conf/cvpr/SalvadorGBD21}, which uses cosine similarity in the common space for ranking the relevant recipes and perform im2recipe retrieval on validation dataset in Recipe 1M \cite{DBLP:conf/cvpr/WangSLLH19}. The performance metrics we report include median rank (MedR) and recall rate at top K (R@K, K = 1, 5, 10) for all the retrievals. We measure the cost by the avg training time per epoch and convergence speed. We want the model to be easy and fast to train because we want our solution to be friendly for other developers to iterate and build upon. Table \ref{tab:SOTA_performance} demonstrates that the hierarchical transformer model in \cite{DBLP:journals/pami/MarinBOHSAWT21} is converging faster: it achieves MedR = 16.2 at second Epoch while the LSTM model needs to complete 20 epochs to achieve a better performance (MedR = 14.15, at Epoch 19 the MedR is 17.9). But it takes the transformer model ~9.5 hours to complete just one epoch, which is very costly for most mobile application development.\\
We need a more efficient network architecture. That's why we decide to move to MobileNets \cite{DBLP:journals/corr/HowardZCKWWAA17}, a family of small, low-latency, low-power neural network models parameterized to meet the resource constraints on mobile device or embedded applications. Specifically, we use MobileNet-v2 \cite{DBLP:conf/cvpr/SandlerHZZC18} in our final model architecture. MobileNet-v2 pushes the state of the art in accuracy vs. latency trade-off for mobile visual recognition tasks, which is a perfect fit for this project. Table \ref{tab:model comparision table} shows that MobileNe-v2 significantly reduces the number of parameters and Floating Point Operations Per Second (FLOPS) compared to the main-stream networks with regular convolutions.

\newcommand{\retrievedindosso}{\fbox{\begin{minipage}{14em}
\tiny
2 cups veal demiglace (16 fl oz)*\\
4 (1-lb) meaty cross-cut veal shanks (osso buco), each tied with kitchen string
1 teaspoon salt\\
1/4 teaspoon black pepper\\
1/4 cup all-purpose flour\\
1 1/2 tablespoons olive oil\\
3 tablespoons unsalted butter\\
2 medium onions, cut into 1/4-inch dice (2 cups)\\
2 medium carrots, cut into 1/4-inch dice (1 cup)\\
2 celery ribs, cut into 1/4-inch dice (1 cup)\\
1 garlic clove, finely chopped\\
1 1/2 cups dry red wine\\
1 (28-oz) can whole tomatoes in juice, drained and coarsely chopped\\
1 turkish or 1/2 california bay leaf (preferably fresh)\\
2 teaspoons chopped fresh flat-leaf parsley\\
1 teaspoon finely grated fresh orange zest\\
3/4 teaspoon finely chopped fresh rosemary\\
3/4 teaspoon finely chopped fresh thyme\\
accompaniment: wild mushroom risotto\\
\end{minipage}}}

\newcommand{\trueindosso}{\fbox{\begin{minipage}{14em}
\tiny
Title: Osso Buco\\
4 ounces pancetta, diced into 1/4 inch cubes (see recipe note)\\
2 1/2 to 3 pounds veal shanks (4 to 6 pieces 2 to 3 inches thick)\\
1/2 cup diced carrot (1/4-inch dice)\\
1/2 cup diced celery (1/4 inch dice)\\
1 medium onion (1/4 inch dice)\\
2 tablespoons chopped garlic (about 4 cloves)\\
3 to 4 sprigs fresh thyme (or 1 teaspoon dried)\\
1 cup dry white wine\\
1 to 2 cups chicken or veal stock\\
Flour for dusting the meat before browning\\
Salt and pepper\\
\end{minipage}}}

\newcommand{\retrievedtitleosso}{\fbox{\begin{minipage}{14em}
\small
1. braised veal shanks\\
2. chipotle lamb chops\\
3. beef short ribs in chipotle and green chili sauce\\
4. haitian pork griot\\
\end{minipage}}}

\newcommand{\retrievedindmatcha}{\fbox{\begin{minipage}{14em}
\tiny
Ingredient: \\
1 medium avocados \\
1 large cucumbers \\
1 each lemon juice of \\
1/2 cup parsley leaves fresh, chopped \\
2 cups ice cubes crushed \\
1 slices lemon or cucumber peel \\
\end{minipage}}}

\newcommand{\trueindmatcha}{\fbox{\begin{minipage}{14em}
\tiny
Title: Iced Matcha Bubble Tea\\
Ingredients \\
1 teaspoon matcha green tea powder\\
2 Tablespoons hot water \\
2 Tablespoons honey, or other sweetener, to taste \\
1 cup (240 ml) Almond Milk \\
½ cup (120 g) ice cubes, plus more for serving \\
¼ cup (50 g) cooked tapioca pearls \\
\end{minipage}}}

\newcommand{\retrievedtitlematcha}{\fbox{\begin{minipage}{14em}
\small
1. avocado aperitif\\
2. super green juice\\
3. pineapple mojito gelatin shot\\
4. kiwi, apple and mint juice recipe\\
5. green smoothie with spinach and greek yogurt\\
\end{minipage}}}

\newcommand{\osso}{\includegraphics[width=6em]{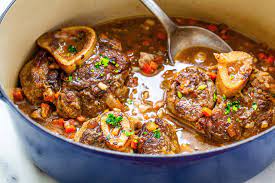}}
\newcommand{\matcha}{\includegraphics[width=6em]{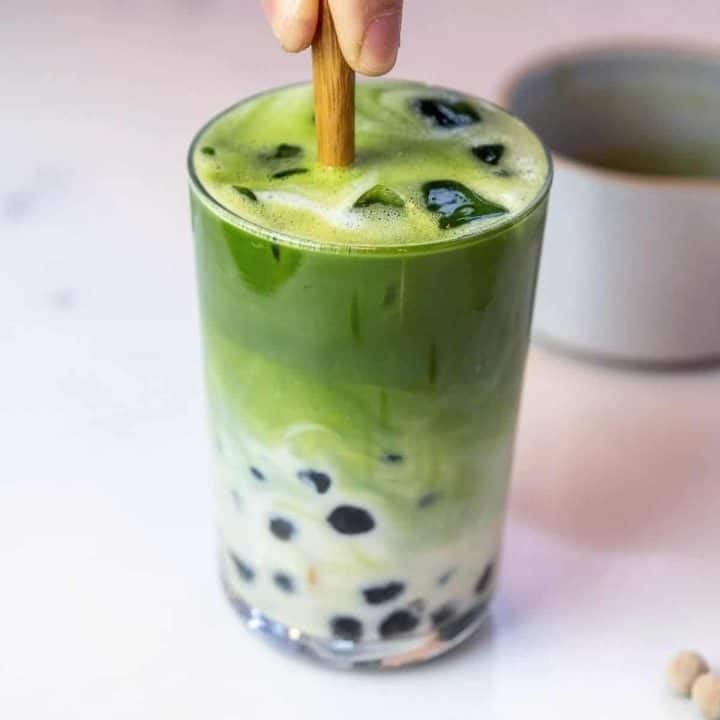}}
\newcolumntype{C}{>{\centering\arraybackslash}m{14em}}
\begin{table*}[t]
\caption{Image2Recipe Retrieval Example}
\label{tab:image query exmple}
\begin{tabular}{l*3{C}@{}}
\toprule
 Query Image & True Ingredients & Retrieved Title & Retrieved Ingredients@1  \\ 
\midrule
\raisebox{-.8\height}\osso & \trueindosso  & \retrievedtitleosso & \retrievedindosso\\ 
\midrule
\raisebox{-.8\height}\matcha & \trueindmatcha  & \retrievedtitlematcha & \retrievedindmatcha\\ 
\bottomrule 
\end{tabular}

\end{table*} 


\section{Experiments and Results}
\subsection{Computation Resource}
Most of our experiments were conducted on Google Cloud Platform. We created a Google Cloud Storage bucket to restore training data and created a Deep Learning VM instance with one NVIDIA Tesla K80 GPU and 2 CPUs to train the model. We followed the \href{https://cloud.google.com/ai-platform/deep-learning-vm/docs/cloud-marketplace}{getting started guide} to set up VM instance.  

\subsection{Reproduce Previous Works}
We started with reproducing results from LSTM model \cite{DBLP:journals/pami/MarinBOHSAWT21} and hierarchical transformer model \cite{DBLP:conf/cvpr/SalvadorGBD21}. It took us some efforts to set up the environment correctly. However, we quickly realized that training from the scratch is too slow and too expensive for us. As Table \ref{tab:SOTA_performance} shows, LSTM model consumes 1 epoch data in 2.5 hours but converges slowly and hierarchical transformer converges fast but is slow in data consumption. Training a model on par with the best performance on the paper would charge us more than one thousand dollars. Hence we stopped reproducing previous works from scratch and decided to utilize the best checkpoint provided by authors. 

\subsection{Build Recipe Library}

As we explained above, we adopted the recipe encoder and best checkpoint from \cite{DBLP:conf/cvpr/SalvadorGBD21}. As Figure \ref{tab:build recipe library} shows, recipes from Recipe1M were passed to recipe encoder to extract embeddings. A embedding vector has a dimension of $1 \times 2014$. All recipe embeddings were stacked into a $N \times 1024$ array, where N is the number of recipes. Using embedding id as key, corresponding recipes were stored in a recipe dictionary . A recipe library consists of a recipe embedding array and a recipe dictionary. We built two individual recipe libraries based on the training and validation partition in Recipe1M dataset. The training library was used to train image encoder and the validation library was used for recipe prediction. As Table \ref{tab:table1} shows, the number of recipes in training partition is about five times of those in validation partition. To reduce memory cost, validation library was adopted for image query. We also provided a training library option for users.    

\subsection{Train Image Encoder}
As we explained in Section \ref{s_approach}, we used MobileNet-V2 as image encoder backbone. Pre-computed recipe embeddings served as ground true labels. In this way, image encoder was forced to learn projecting a image into an embedding close to recipe embedding (Figure \ref{tab:image encoder training}). MobileNet-V2 was initialized with the parameters pre-trained on ImageNet. In this way, we succeeded to train a converged image encoder in 10 hours with MedR=2.0 (\ref{tab:SOTA_performance}).   

We used bi-directional triplet loss function for model training. We only briefly introduce the loss function here. Readers can check \cite{DBLP:conf/cvpr/SalvadorGBD21} for more details. A triplet loss function can be written as follows: 
\begin{equation}
     L_{cos}(a, p, n) = max(0, c(a, n) - c(a, p) + m
\end{equation}
where a, p, and n refer to the anchor, positive, and negative samples, c(.) is the cosine similarity metric, and m is the margin (m=0.3 in our work). 
A bi-directional triplet loss function on feature sets a and b can be defined as:
\begin{equation}
\begin{split}
     L_{bi}^{'}(i,j) &= L_{cos}(a^{n=i}, b^{n=i}, b^{n=j}) \\
                    &+ L_{cos}(b^{n=i}, a^{n=i}, a^{n=j})\\
\end{split}
\end{equation}

\subsection{Predict Recipe From Image}
As Figure \ref{tab:predict recipe from image} shows, the process is quite straightforward: 1) convert query images into image embedding with image encoder, 2) compute the cosine similarity with recipe embedding array, 3) retrieve top k (here we use k=5) recipe ids, and 4)query top k recipes from recipe dictionary with recipe ids. We list some image query results in Table \ref{tab:image query exmple}. The retrieved recipe is very close to what we queried.  

\subsection{Expand Recipe Library}

We created a \href{https://github.com/jianfa/RecipeSnap-a-lightweight-image-to-recipe-model}{RecipeSnap} class integrating all important modules, such as image encoder loading, recipe encoder loading, image preprocessing, recipe preprocessing, recipe prediction, and recipe library update. With this class object, users can easily query recipes from certain image or add new recipe to the recipe library. 

Collecting all recipes was an impossible mission. Our model makes this task feasible. Smart phone is accessible to almost everyone. If our model can be deployed on smart phones, it would be very easy for everyone to contribute new recipes to our recipe library.

\section{Discussion}

People are looking for easier ways to keep cooking fit into their busy schedules in today's fast-paced life. As a result, many meal-planning services are booming. Those services try to simplify the cooking by planning everything for their customers ahead. But people who use these services may get stuck with things they don't like. We believe it's essential to keep the joy for people to discover attractive dishes by themselves and let them feel the fulfillment from cooking beautiful and tasty food. So we want to build this application where people can customize their recipe library and easily query the recipe and ingredients from their phone with any dish images they like. 

In the future, we plan add more features to advance this purpose. For example, people usually face the dilemma of seeing a few ingredients left in the fridge and wondering how they can utilize them. An ingredients-to-recipe function can be quite applicable for this use case. Beyond that, people may want the feature to rank or filter the queried recipe, given their nutrition and budget requirements. We can also add a feedback module to let the application learn the user's preference so that the query results can be more intelligent. To sum up, many interesting features can be added to what we have built now, and we believe they will benefit foodies who are busy but still pursue the fun of cooking.

\section{Conclusion}
In this paper, we introduce RecipeSnap, a lightweight image-to-recipe retrieve model developed to handle the the problem of automation for recognition of images of cooking dishes and finding the relevant recipes containing information including ingredient lists, instructions etc.. The experiments conducted using multiple existing solutions and RecipeSnap imply that our model design help reduce memory cost and computational expense for model training and implementing while obtain a compatible performance comparing to current state-of-art neural networks. The use of the pre-trained recipe encoder as part of the recipe encoder and the cooked recipe library collected using recipes from a large scaled dataset Recipe1M make RecipeSnap easier and faster to train and deploy. We adopt MobileNet-V2 instead of other deeper complex networks as the backbone for image encoder, which makes RecipeSnap suitable to portable devices and platforms like smart phones and tablets. 

This model can easily be developed into applications for smart phone and tablets where potential users can customize their recipe library and easily query the recipes from their phones with any dish images they are interested at. In the future, we plan to keep contribute to this tool by exploring more meaningful features to advance this purpose.

\begin{table*}
\begin{center}
\caption{Contributions of team members.}
\label{tab:contributions}
\begin{tabular}{|l|c|p{8cm}|}
\hline
Individual & Contributed Aspects & Details \\
\hline\hline
Jianfa Chen & \makecell{Propose idea, model training,\\ implementation, and writing}  & Propose the idea of lightweight model. Build recipe library. Train image encoder with MobileNet V2 backbone. Major contributor to RecipeSnap code. Contribute to the writing of abstract, experiments and results, and conclusion section. \\
Yue Yin & \makecell{Model implementation and experimentation\\ Propose extension ideas and writing} & Run experimentation to train and compare SOTA models. Implement the image pre-processing module. Propose ideas for future extension. Contribute to the writing of Approach and Discussion.  \\
Yifan Xu & \makecell{Implementation, Analysis and Reporting} & Explore and study related works in the specific field. Experiment with alternative ways of model implementation and deployment. Implement recipe preprocessing module. Conduct unit testing with model code. Contribute to the introduction and conclusion sections of the report. \\

\hline
\end{tabular}
\end{center}
\end{table*}



\section{Work Division}

Individual contributions are summarized in Table \ref{tab:contributions}.

{\small
\bibliographystyle{ieee_fullname}
\bibliography{recipesnap}
}

\end{document}